# MESA: Maximum Entropy by Simulated Annealing

**Gerhard Paaß**
German National Research Center for Computer Science (GMD)
Schloß Birlinghoven
D-5205 St.Augustin

## Abstract

Probabilistic reasoning systems combine different probabilistic rules and probabilistic facts to arrive at the desired probability values of consequences. In this paper we describe the MESA-algorithm (Maximum Entropy by Simulated Annealing) that derives a joint distribution of variables or propositions. It takes into account the reliability of probability values and can resolve conflicts between contradictory statements. The joint distribution is represented in terms of marginal distributions and therefore allows to process large inference networks and to determine desired probability values with high precision. The procedure derives a maximum entropy distribution subject to the given constraints. It can be applied to inference networks of arbitrary topology and may be extended into a number of directions.

## 1 INTRODUCTION

Probabilistic inference networks [Pearl 88] have been used to model uncertain relations between variables, for instance in the area of medical diagnosis. They represent the information contained in probabilistic rules and the information on the probability of certain facts. This induces a joint probability distribution, usually for a large set of variables.

In case of more complex probability networks, the dependency graph between variables may contain cycles. In this situation the classical update mechanisms [Lauritzen & Spiegelhalter 88] for point probabilities cannot work. In addition the probabilities supplied by the experts may be *incoherent*, such that there is no common probability measure that simultaneously meets all constraints. Hence we consider the elicitation of probabilities as some sort of *measurement process* which may be disturbed by 'measurement noise'. In this paper we use the maximum likelihood approach to derive a joint probability distribution that is an optimal compromise between the different supplied probabilities. Each probability influences this compromise according to its reliability (i.e. variance of measurement noise). Higher order interactions not affected by the observed probabilities are set to zero according to the maximum entropy principle. The resulting distribution is a best 'guess' in the sense that it contains least statistical information from all distributions with optimal fit to the marginal constraints.

In earlier papers [Paaß 89, Paaß 91, Paaß 91a] we have proposed to generate a synthetic sample by the simulated annealing algorithm which represents the joint distribution. This sample only approximates the theoretical maximum entropy distribution as it consists of a finite number $n$ of elements and the probability values are restricted to the numbers $0/n, 1/n, \ldots n/n$. Consequently the procedure was unable to determine conditional probabilities with respect to rare events, which seriously limited its applicability to real world problems.

In this paper we extend this algorithm to the case of continuous probability values. We do no longer construct a joint synthetic sample, but update *marginal* distributions, each of which corresponds to a supplied rule or fact. Therefore no longer complete synthetic samples with a large number of records but only the relatively small probability vectors of marginal distributions have to be stored. The resulting procedure is called **MESA**-algorithm (Maximum Entropy by Simulated Annealing) and allows to determine the maximum entropy solution to an arbitrary precision. It can handle cyclic dependencies of rules and takes into account the relative reliability of probability statements. It can be extended into various directions, e.g. nonlinear constraints, upper and lower probabilities, default reasoning, Bayesian second order probabilities, continuous variables, etc (cf. [Paaß 90]). The MESA-algorithm has been implemented in Common Lisp on a Sun Sparcstation.

In the next section we introduce the basic notation and cost functions measuring the reliability of probability statements. The following two sections discuss the convergence properties of the simulated an-nealing algorithm and show that under specific conditions it generates a synthetic sample that optimally fits the constraints and in addition has maximum entropy within the class of all distributions with optimal fit. The subsequent section discusses marginal models as an alternative representation. In the following section we formulates the algorithm in terms of the marginal distributions. The last section contains a summary.

## 2 NOTATION AND COST FUNCTIONS

Assume there is a vector $x := (x_1, \ldots, x_k)'$ of binary random variables $x_i$. Then $x$ may take $m := 2^k$ different values forming the sample space $\mathcal{X} = \{\xi_l \mid l = 1, \ldots, m\}$. Corresponding to each $x_i$, a proposition $A_i$ is defined which holds if and only if $x_i = 1$. If $\mathcal{B}$ is the Boolean algebra generated from the $A_i$, each proposition $B \in \mathcal{B}$ corresponds to a subset of $\mathcal{X}_B \subset \mathcal{X}$. To arrive at a simpler notation we write $\xi \in B$ instead of $\xi \in \mathcal{X}_B$. Let $p(x)$ denote the common unknown distribution of the $x_i$, and let $q := (q_1, \ldots, q_m)$ with $q_l := p(x=\xi_l)$, $\xi_l \in \mathcal{X}$, be the vector of unknown parameters of $p(x)$. Then $\mathcal{P} := \{p \in \Re^m \mid p_i \geq 0; \sum_{i=1}^m p_i = 1\}$ is the set of all possible probability vectors.

We consider a *probabilistic inference network* where the expert's knowledge is stated in terms of "probabilistic rules" which may be interpreted as restrictions on conditional probabilities

$$p(D^j \mid B^j) = t^j \qquad j = 1, \ldots, s \qquad (1)$$

for propositions $D^j, B^j \in \mathcal{B}$. The *influence set* $\mathcal{I}^j$ is defined as the set of variable indices which enter the definition of the rule. An example for a probabilistic rule is $p(A_9 \wedge A_5 \mid A_3 \vee A_4) = 0.1$ with an associated influence set $\mathcal{I}^j := \{3, 4, 5, 9\}$. We define $\mathcal{I} := \{\mathcal{I}^1, \ldots, \mathcal{I}^s\}$ as the set of all influence sets. Marginal probabilities, e.g. $p(A_1 \vee A_5) = 0.8$, are a special case of probabilistic rules, as they are conditional probabilities with respect to the tautology.

In practical applications experts usually have only a vague idea of the actual probability values $t^j$. This is taken into account by assuming that each $t^j$ is determined from an independent sample $S^j$ containing $n^j$ elements. The observed statistics (e.g. counts) of samples may differ from their theoretical values because of the sampling error. This error decreases with growing sample size $n^j$. This means that the observed values $\tilde{t}^j$ may be *contradictory* to some extent and corresponds to the actual experience that the rules specified by experts may be inconsistent. Hence there may exist no $p(x)$ with parameter $q$ such that $\tilde{t} = t(q)$ for $\tilde{t} := (\tilde{t}^1, \ldots, \tilde{t}^s)$ and $t(q) := (t^1(q), \ldots, t^s(q))$. To measure the deviation between $\tilde{t}$ and $t(q)$ we use a real-valued *cost function* $C(t(q))$ and we would like to find a joint distribution with minimal cost.

Let $p(t^j \mid q)$ be the conditional distribution of the available statistic $t^j$, if $q$ is the probability vector of the underlying distribution. If we consider $t$ as given and $q$ as variable this defines the *likelihood function*

$$p(t \mid q) := \prod_{j=1}^s p(t^j \mid q)$$

According to the maximum likelihood approach the 'cost' of a parameter $q$ is minimal if its likelihood $p(t \mid q)$ is maximal. Therefore we can define a *log-likelihood cost function*

$$C(q) := \sum_{j=1}^s C^j(q) \qquad C^j(q) := -\log p(\tilde{t}^j \mid q) \quad (2)$$

If $t^j$ is a marginal distribution with $g$ groups, observed counts $t_1^j, \ldots, t_g^j$ and underlying probabilities $p_1^j(q), \ldots, p_g^j(q)$, we have a multinomial distribution with loglikelihood function [Bishop et al.75, p.65]

$$C^j(q) = -\sum_{i=1}^g t_i^j \log \left( n^j p_i^j(q) \right) \qquad (3)$$

For a probabilistic rule the sample is 'truncated', i.e. the counts in the $g$-th group are unknown. Then the likelihood is formed with respect to the conditional probabilities $p_i^j(q)/\sum_{l=1}^{g-1} p_l^j(q)$. In [Paaß 90] the relation of this expression to the Kullback-Divergence [Gokhale & Kullback 78, p.38] and in [Paaß 91a] the extension to Bayesian prior distributions is discussed.

## 3 GENERATING A JOINT SYNTHETIC SAMPLE

We may use a stochastic optimization procedure to generate a *synthetic* sample $X = (x(1), \ldots, x(n)) \in \mathcal{X}^n$ of size $n$ which optimally fits to the constraints (1). Let $\mathcal{Q}_n := \{q(X) \in \mathcal{P} \mid X \in \mathcal{X}^n\}$ be the corresponding set of 'empirical' distributions. Because of the law of large numbers the empirical parameters of a random sample of the true distribution $p(x)$ with growing sample size converge to the true parameters of the distribution. Hence there *exist* samples $X$ with parameters that are arbitrarily close to the true parameters if $n$ is taken large enough.

The construction of an optimal sample $X^*$ with minimal cost $C^* := C(X^*)$ may be performed by the *simulated annealing* algorithm [Aarts & Laarhoven 85] [Aarts & Korst 88] which converges to a global optimum and proceeds in the following way:

1. Select an arbitrary starting value for $X$.

2. With probability $P^{mod}(X_j \mid X_i)$ randomly change the present sample $X_i$ to a 'modified' sample $X_j \in \mathcal{X}^n$.

3. Accept the modification with probability $P_\beta^{acc}(j \mid i) = 1$ if $C(X_j) < C(X_i)$ and

$$P_\beta^{acc}(j \mid i) = \exp\left([C(X_i) - C(X_j)]\beta\right) \qquad (4)$$

otherwise. If accepted, $X_j$ becomes the present sample; otherwise the old sample $X_i$ is kept. Continue with step 2.

After a number of iterations the parameter $\beta$ is increased to $\beta := \gamma\beta$ (a typical value for $\gamma$ is 1.1). The procedure stops after a number of cycles. [Aarts & Laarhoven 87, p.215ff] discuss other schedules for increasing $\beta$ and more advanced stopping criteria. Assume the acceptance probabilities are defined according to (4) and the following two conditions hold:

$$P^{mod}(X_j \mid X_i) = P^{mod}(X_i \mid X_j) \qquad (5)$$

for all $X_i, X_j \in \mathcal{X}^n$.

Each $X_i \in \mathcal{X}^n$ can be modified to any other $X_j \in \mathcal{X}^n$ with positive probability in a finite number of steps. (6)

Then [Aarts & Laarhoven 85, p.201] for fixed $\beta$ the Markov chain of successive synthetic samples $X_i$ converges to an equilibrium distribution $\Pr_\beta$ which is independent of the starting state

$$\Pr_\beta(X) = c_0(\beta)\exp(-C(X)\beta) \qquad (7)$$

The constant $c_0(\beta) = 1/\sum_{X\in\mathcal{X}^n} \Pr_\beta(X)$ normalizes the sum of probabilities to one. For $\beta \to \infty$ the probability of accepting a modification approaches zero except for modifications yielding samples with lower cost values. The corresponding limit distribution concentrates on the set of samples with minimal cost which form the set of *global minima*. With appropriate schedules for increasing $\beta$ the execution time of the simulated annealing algorithm is proportional to a polynomial in the system complexity (number of variables and restrictions) [Aarts & Laarhoven 85, p.216], [Mitra et al.86].

We may use a simple procedure to generate a modification:

First randomly select (with equal probability) one or more records $x(i)$ in $X$. Then randomly select (with equal probability) some components of those records and change their values. (8)

For $C(X) = const$ relation (7) yields the stationary distribution $\Pr^\circ(X) = c_0$, which is independent of $\beta$. As $\sum_{X\in\mathcal{X}} c_0 = 1$ we calculate $c_0 = 1/\#(\mathcal{X}^n)$ as the inverse of the number of samples in $\mathcal{X}^n$. Therefore a probability vector $q$ is generated with a probability $\Pr^\circ(q) = \sum_{X\in\mathcal{X};q(X)=q} c_0 = N_n(q)/\#(\mathcal{X}^n)$ which is proportional to the number $N_n(q)$ of different samples $X$ with identical counts $q$.

Using (7) we get for the loglikelihood cost function $C(X) = -\log p(\tilde{t} \mid q(X))$ and fixed $\beta > 0$ the stationary distribution

$$\begin{aligned}\Pr_\beta(X) &= c_0(\beta)\exp(\beta\log p(\tilde{t} \mid q(X))) \\ &= c_1(\beta)p(\tilde{t} \mid q(X))^\beta\end{aligned}$$

with a constant $c_1(\beta)$. As $\Pr_\beta(X)$ is a probability we have $\sum_{X\in\mathcal{X}} c_1(\beta)p(\tilde{t} \mid q(X))^\beta = 1$. Then

$$\begin{aligned}\Pr_\beta(q) &= \sum_{X;q(X)=q} c_1(\beta)p(\tilde{t}\mid q)^\beta \\ &= \frac{\sum_{X;q(X)=q} c_1(\beta)p(\tilde{t}\mid q)^\beta}{\sum_{r\in\mathcal{Q}}\sum_{X;q(X)=r} c_1(\beta)p(\tilde{t}\mid r)^\beta} \\ &= \frac{N_n(q)p(\tilde{t}\mid q)^\beta}{\sum_{r\in\mathcal{Q}} N_n(r)p(\tilde{t}\mid r)^\beta} \\ &= \frac{[N_n(q)/\#(\mathcal{X}^n)]p(\tilde{t}\mid q)^\beta}{\sum_{r\in\mathcal{Q}}[N_n(r)/\#(\mathcal{X}^n)]p(\tilde{t}\mid r)^\beta} \\ &= \frac{\Pr^\circ(q)p(\tilde{t}\mid q)^\beta}{\sum_{r\in\mathcal{Q}}\Pr^\circ(r)p(\tilde{t}\mid r)^\beta}\end{aligned}$$

Therefore $\Pr_\beta(q)$ can be considered as the posterior distribution that would result in a Bayesian analysis, if we take $\Pr^\circ(q)$ as prior distribution and have likelihoods $p(\tilde{t}\mid q)^\beta$. The term

$$p(\tilde{t}\mid q)^\beta = p(\tilde{t}\mid q)\cdots p(\tilde{t}\mid q)$$

can be interpreted as a likelihood function where the same data $\tilde{t}$ independently has been observed $\beta$ times. Hence increasing $\beta$ to infinity is equivalent to calculating the Bayesian posterior function for the case that the number of observations approaches infinity.

Suppose the cost function is constant $C(X) = const$. Let us for a moment consider the $x(i)$, $i = 1,\ldots,n$ as i.i.d. discrete random variables with the values $\xi \in \mathcal{X}$. If we use a *noninformative distribution* with $P(x(i){=}\xi) = 1/m$, then every possible sequence $(\xi_1,\ldots,\xi_n)$ of values – and hence the corresponding sample $X \in \mathcal{X}^n$ – has identical probability and therefore

$\Pr^\circ(q)$ is the sampling distribution of the non-informative multinomial. (9)

Consequently the empirical frequencies $nq(X)$ follow the corresponding multinomial distribution with expected values $n/m$, variances $(1-1/m)/(mn)$ and covariances $-1/(nm^2)$.

## 4   MAXIMIZING THE ENTROPY OF A SYNTHETIC SAMPLE

Usually the number $s$ of constraints (1) is much smaller than the number of parameters $m$ of the probability distributions $q$. Then there will be multiple solutions with minimal cost. According to the maximum entropy principle [Csiszár 85] it is sensible to select from these solutions one with maximum entropy. The entropy of a discrete distribution $q \in \mathcal{P}$ is defined as

$$H(q) := -\sum_{i=1}^{m} q_i \log q_i \qquad (10)$$

Given non-contradictory constraints the maximum entropy distribution "agrees with what is known, but expresses a 'maximum uncertainty' with respect to all other matters" [Jaynes 68, p.231]. The principle is usually justified on the basis of entropy's unique properties as an information measure. Shore and Johnson [Shore & Johnson 80] require that methods of inductive inference satisfy reasonable consistency axioms. If the consistency axioms are accepted and if information is given in the form of constraints on expected values they prove that only the maximum entropy distribution has to be selected. The following theorem shows that with $n \to \infty$ the simulated annealing algorithm for joint synthetic samples $X$ yields a synthetic sample with cost and entropy simultaneously arbitrarily close to the optimal values.

**Theorem 1**
For the samples generated according to (8) and the algorithm described in the previous section we assume that the prior distribution $\Pr^\circ(X)$ of each sample $X$ is identical and the following conditions hold

$$C^* = \inf_{p \in \mathcal{P}} C(p) \text{ exists and for some } \delta_0 > 0 \text{ the function } C(p) \text{ is continuous in } \quad (11)$$
$$D_{\delta_0} := \{p \in \mathcal{P} \mid C(p) \leq C^* + \delta_0\}.$$

There exist $a_C, b_C > 0$ such that for all $p, q \in D_{\delta_0}$ the cost differences can be bounded by a polynomial

$$\mid C(p) - C(q) \mid \leq a_C \left(\max_i \mid p_i - q_i \mid\right)^{b_C} \quad (12)$$

Let $b_\beta \in (1, 1 + b_C)$ be a constant and $\beta_n := a_\beta n^{b_\beta}$ for some $a_\beta > 0$. Let $\Pr_n$ be the stationary distribution on the elements of $Q_n$ if the algorithm is applied to the samples $X$ in $\mathcal{X}_n$ with the cost function

$$\tilde{C}_n(q(X)) := \beta_n C(q(X)) \quad (13)$$

Then $H(p)$ has a maximum value $H^*$ in $D_0$. In addition for all $\epsilon > 0$ there is a $n_\epsilon$ such that for all $n \geq n_\epsilon$

$$\begin{aligned}\Pr_n(H(q) < H^* - \epsilon) &\leq \epsilon \\ \Pr_n(C(q) > C^* + \epsilon) &\leq \epsilon\end{aligned} \quad (14)$$

If in addition for some $\delta > 0$ the cost function $C$ is continuously differentiable in $D_\delta$ then (14) holds with

$$1 < b_\beta < 2 \quad (15)$$

The proof is given in the appendix. The upper bound on $n$ is necessary to ensure some fluctuations around the expected values of $\Pr^\circ(q)$.

## 5 MARGINAL MODELS

The multinomial distribution $p(x)$ can be parametrized in a different way, which is especially interesting in relation to the maximum entropy property. We may define *cross-product ratios* [Bishop et al.75, p.13ff]

$$\alpha_{\{i\}} := \log \frac{p_i^1}{p_i^0} \qquad \alpha_{\{i,j\}} := \log \frac{p_{ij}^{01} p_{ij}^{10}}{p_{ij}^{11} p_{ij}^{00}} \quad (16)$$

with $p_i^l := p(x_i{=}l)$ and $p_{ij}^{lm} := p(x_i{=}l, x_j{=}m)$. The parameters $\alpha_{\{i\}}, \alpha_{\{i,j\}}$ completely define the joint distribution of $x_{\{i,j\}} := (x_i, x_j)$. Moreover if $x_i$ and $x_j$ are independent then $\alpha_{\{i,j\}} = 0$. Therefore $\alpha_{\{i,j\}}$ measures the interaction between $x_i$ and $x_j$. This can be extended to more variables, e.g. three variables $x_{\{i,j,k\}} := (x_i, x_j, x_k)$

$$\alpha_{\{i,j,k\}} := \log \frac{p_{ijk}^{111} p_{ijk}^{100} p_{ijk}^{010} p_{ijk}^{001}}{p_{ijk}^{000} p_{ijk}^{110} p_{ijk}^{101} p_{ijk}^{011}} \quad (17)$$

with $p_{ijk}^{lmn} := p(x_i{=}l, x_j{=}m, x_k{=}n)$. Generally the numerator contains all probabilities with an uneven number of ones, while the other probabilities are in the denominator. For each subset $A \subseteq \{1, \ldots, k\}$ there is a cross-product ratio $\alpha_A$ (if we define $\alpha_\emptyset := 0$). These $2^k$ parameters $\alpha := (\alpha_\emptyset, \alpha_{\{1\}}, \ldots, \alpha_{\{1,\ldots,k\}})$ are an alternative parametrization of the distribution $p(x)$. As probabilities of 0 lead to infinite logarithms, only probability measures with strictly positive probabilities $q \in \mathcal{P}^+ := \{p \in \Re^m \mid p_i > 0; \sum_i p_i = 1\}$ may be expressed. A discussion of such *marginal models* is given in [Liang et al. 92].

We are interested in the values of $\alpha_A$ for the maximum entropy distribution with respect to the constraints (1). The influence sets $\mathcal{I}^j$ indicate which variables are direcly involved in the $j$-th rule. Obviously cross-product ratios $\alpha_A$ with $A \subseteq \mathcal{I}^j$ can be affected by the $j$-th rule.

Let $\mathcal{K} := \{A \mid \exists_j A \subseteq \mathcal{I}^j\}$ and $\bar{\mathcal{K}} := \{A \subseteq \{1, \ldots k\}\} \setminus \mathcal{K}$. Then in the maximum entropy distribution subject to the constraints (1) each cross-product ratio $\alpha_A, A \in \bar{\mathcal{K}}$ will be zero. For the proof assume that $p^*$ is the maximum entropy distribution with an associated entropy value $H(p^*) = H^*$. If some margins, say margin $j$, involve probabilistic rules we may take the *complete* marginal distribution with respect to the influence set $\mathcal{I}^j$ from $p^*$ and determine the maximum entropy distribution with respect to constraints involving those *complete* margins. Obviously we will get the same optimal distribution $p^*$ as it is unique. [Gokhale & Kullback 78, p.214f] show that the maximum entropy distribution may be calculated by the iterative proportional fitting algorithm (IPF) starting with a distribution where all cross-product ratios are set to 0. The IPF, however, does only change cross-product ratios $\alpha_A, A \in \mathcal{K}$, while the others are not altered [Bishop et al.75, p.98]. Therefore all $\alpha_A, A \in \bar{\mathcal{K}}$ are zero for $p^*$.

## 6 ALGORITHM FOR MARGINAL MODELS

A marginal distribution with respect to $x_B$ is completely determined by the cross-product ratios $\alpha_A$, $A \subseteq B$. If we change some $\alpha_C$ with $C \not\subseteq B$ the marginal $x_B$ is not affected. Therefore we can perform the determination of the maximum entropy distribution $p^*$ in the following way:

1. Randomly modify sample $X_i$ to sample $X_j$ according to (8).

2. Determine the distribution $p_{X_j}$ corresponding to $X_j$ and set all cross-product ratios $\alpha_A$, $A \in \bar{\mathcal{K}}$, to zero yielding a probability distribution $p^+_{X_j}$. Note that the marginals $x_B$, $B \in \mathcal{K}$ are not affected by this change.

3. Determine the cost $C(X_j)$ from the marginals of $p^+_{X_j}$ (which are identical to the marginals of $X_j$) and accept $X_j$ according to (4). Then proceed with 1.

Because the cost only depends on the marginals, we no longer need the complete sample $X$ but may formulate the algorithm completely in terms of the margins $x_{\mathcal{I}^j}$, $j = 1, \ldots, s$. As we only have to store a limited number of marginal probabilities and not the values of individual records, the size $n$ of the synthetic sample may be arbitrarily large and the approximation of the maximum entropy distribution will become arbitrarily good.

We have to define a modification procedure for synthetic samples $X_i$ which obeys the conditions (5) and (6), and for constant cost function generates a sample distributed according to the non-informative distribution $\Pr^\circ(X)$. We may assume that the sample size $n$ of $X_i$ is arbitrarily large and that $\alpha_A \approx 0$ for all $A \in \bar{\mathcal{K}}$ with an arbitrarily small deviation. Therefore step 1. above, the modification of $X_i$, my be decomposed in the following way:

1.1 Randomly select some marginal $x_{\mathcal{I}^j}$.

1.2 Setup a sample $X_{\mathcal{I}^j}$ of size $n_a$ containing only the variables $x_{\mathcal{I}^j}$, such that the distribution of $X_{\mathcal{I}^j}$ is equal to the marginal $x_{\mathcal{I}^j}$-distribution of $X_i$.

1.3 Independently modify randomly chosen variables of randomly selected records in $X_{\mathcal{I}^j}$, the probability of values 0 and 1 being equal to 0.5.

1.4 Determine the cross product ratios $\alpha_A$, $A \subseteq \mathcal{I}^j$, from $X_{\mathcal{I}^j}$.

1.5 Modify the original sample $X_i$ in such a way that its cross-product ratios $\alpha_A$, $A \subseteq \mathcal{I}^j$, take the values determined for $X_{\mathcal{I}^j}$, while the remaining cross-product ratios remain constant. This yields the modified sample $X_j$.

Obviously the procedure has the necessary properties symmetry (5) and reachability (6) and generates the non-informative prior $\Pr^\circ(X)$ except that all $\alpha_A$, $A \in \bar{\mathcal{K}}$, will have already the 'optimal' value 0 (with infinitesimal deviations).

Because $X_i$ is a sample, its adaption to the new cross-product ratios can be done only approximately as only probability values $l/n$ with integer $l$ are feasible. However, as $n$ may be arbitrarily large, we may approximate all $\alpha$-s arbitrarily well. As the cost function depends only on the margins $x_{\mathcal{I}^j}$, $\mathcal{I}^j \in \mathcal{I}$, we can ignore the joint sample $X_i$ completely and use only the exact marginal distributions $x_{\mathcal{I}^j}$. Then the discretization problem vanishes. This yields the final MESA-procedure:

I.) Initialize the distribution of the margins $x_{\mathcal{I}^j}$, $\mathcal{I}^j \in \mathcal{I}$, in such a way that they are compatible, i.e. the common cross-product ratios $\alpha_A$, $A \in \mathcal{K}$, have identical values for the different margins.

II.) Randomly select some marginal $x_{\mathcal{I}^j}$.

III.) Modify the probabilities of $x_{\mathcal{I}^j}$ such that their joint distribution corresponds to the distribution of probabilities in a sample $X_{\mathcal{I}^j}$ of the variables $x_{\mathcal{I}^j}$ and of size $n_a$ with non-informative distribution. As only the asymptotic distribution is relevant the distribution may be modified in such a way that the discrete histogram resulting for finite samples is continuously interpolated.

IV.) Determine the cross product ratios $\alpha_A$, $A \subseteq \mathcal{I}^j$, from $X_{\mathcal{I}^j}$.

V.) Adapt the remaining marginal distributions $x_{\mathcal{I}^r}$, $\mathcal{I}^r \in \mathcal{I} \setminus \{\mathcal{I}^j\}$ such that they get the cross product ratios $\alpha_A$, $A \subseteq \mathcal{I}^j$. The other cross-product ratios remain constant. Together all margins define the modified probability distribution $p^+$.

VI.) Determine the cost $C(p^+)$ from the marginals $x_{\mathcal{I}^j}$, $\mathcal{I}^j \in \mathcal{I}$, and accept the modifications according to (4). Then proceed with step II.).

The fictitious sample size $n_a$ determines the variance of the non-informative distribution around its expected values. As we have differentiable cost functions, according (15) the annealing parameter $\beta$ should be increased as a function of $n_a$ $\beta := a_\beta n_a^{b_\beta}$ for some $a_\beta > 0$ and $1 < b_\beta < 2$. Then the simulated annealing procedure will yield the marginals of the maximum entropy distribution subject to the constraints (1).

The adaption in step V.) may be done directly in terms of the cross-product ratios. This amounts to a nonlinear equation system which analytically may be solved for margins involving up to four variables. For larger marginals iterative methods for the solution of nonlinear equation systems may be employed. Alternatively we might use the iterative proportional fitting (IPF) algorithm to adapt the margins $x_{\mathcal{I}^r}$ to the marginals of $x_{\mathcal{I}^j}$. This approach has the advantage that it also works for distributions with zero probabilities, where the cross-product ratios take the value infinity. In the actual implementation of the MESA-algorithm the IPF is employed.

In the case of diagnosis we are mainly interested in the conditional distribution of some variables ('diseases') conditional to specific values of other variables ('symptoms'), e.g. the conditional distribution $p(A_1 \lor A_2 \mid A_7 \land A_8)$. Then we simply may setup an additional margin $x_{\{1,2,7,8\}}$ and include it into the cal-culations in the same way as the other margins $x_{\mathcal{I}^j}$, $\mathcal{I}^j \in \mathcal{I}$. The only exception is that there is no observation or rule available for this margin. Then the algorithm will generate the joint distribution of the variables $x_1, x_2, x_7, x_8$ and we can calculate the desired conditional distribution $p(A_1 \vee A_2 \mid A_7 \wedge A_8)$.

## 7 SUMMARY

The MESA-algorithm constructs a joint distribution of variables, that is compatible in an optimal way with the given probabilistic facts and rules. It uses the maximum likelihood criterion to resolve conflicts between the constraints taking into account the relative reliability of the constraints. Within the set of cost-optimal solutions a distribution with maximum entropy is selected. This ensures that higher order interactions between variables are zero unless there is explicit information on a dependency.

The algorithm can be formulated in terms of the marginal distributions each of which corresponds to a single probabilistic rule or fact. Because of these moderate storage requirements large inference networks can be processed. It is based on a general global stochastic optimization procedure which has shown good performance even in difficult problems. The optimization consists of successive changes of marginal probabilities ensuring that all marginals remain consistent. The procedure can be considered as a general constraint relaxation mechanism and therefore can be applied to a large number of reasoning tasks.

The MESA-algorithm has been implemented on a Sparcstation and shows promising results. In contrast to earlier versions it is not hampered by the discreteness of synthetic samples and can determine results to an arbitrary precision without excessive requirements of storage an computational effort. During the next time we will test the algorithm with realistic inference networks to investigate its properties.

### Acknowledgements

This work was supported by the German Federal Department of Research and Technology, grant ITW8900A7.

## References

[Aarts & Korst 88] Aarts, E., Korst, J. (1988): *Simulated Annealing and Boltzmann Machines.* Wiley, Chichester

[Aarts & Laarhoven 85] Aarts, E.H.L., van Laarhoven, P.J.M. (1985), *Statistical Cooling: A General Approach to Combinatorial Optimization Problems*, Philips J. Res., Vol.40, pp193-226

[Aarts & Laarhoven 87] Aarts, E.H.L., van Laarhoven, P.J.M. (1987), *Simulated Annealing: A Pedestrian Review of the Theory and some Applications*, in: Devijver, P.A., Kittler, J. (eds.) *Pattern Recognition Theory and Applications*, Springer-Verlag, Berlin

[Bishop et al.75] Bishop, Y.M.M., Fienberg, S.E., Holland, P.W. (1975), *Discrete Multivariate Analysis: Theory and practice*, MIT-Press, Cambridge, Mass.

[Csiszár 85] Csiszár, I. (1985), An Extended Maximum Entropy Principle and a Bayesian Justification. In: J.M. Bernard, M.H. DeGroot, D.V. Lindley, A.F.M. Smith (eds.): Bayesian Statistics 2, North Holland, Amsterdam, pp.83-98.

[Csiszár & Körner 81] Csiszár, I., Körner, J. (1981), Information Theory. Coding Theorems for Discrete Memoryless Systems, Academic Press, New York

[Darroch & Ratcliff 72] Darroch, J.N., Ratcliff, D. (1972): Generalized Iterative Scaling for Log-Linear Models. *The Annals of Mathematical Statistics*, Vol.43, p.1470-1480

[Gokhale & Kullback 78] Gokhale, D.V., Kullback, S. (1978): *The Information in Contingency Tables,* Marcel Dekker, New York

[Green & Heller 81] Green J., Heller, W.P. (1981), Mathematical Analysis and Convexity with Applications to Economics, in: Arrow, K.J., Intriligator, M.D. (eds) Handbook of Mathematical Economics, North Holland, Amsterdam

[Jaynes 68] Jaynes, E.T. (1968), Prior Probabilities, IEEE-Transactions on Systems Science and Cybernetics, SSC-4, 227-241

[Lauritzen & Spiegelhalter 88] Lauritzen, S.L., Spiegelhalter, D.J. (1988): Local Computations with Probabilities on Graphical Structures and their Application to Expert Systems, *J. Royal Statistical Soc. Ser. B*, Vol. 50, pp.157-224

[Liang et al. 92] Liang, K., Qaqish, B., Zeger, S.L. (1992): MultivariateRegression Analyses for Categorical Data (with discussion). *J. Royal Statistical Soc. Ser. B*, Vol.54, pp.3-40.

[Mitra et al.86] Mitra, D., Romeo, F., Sangiovanni-Vincentelli, A. (1986): Convergence and finite time behaviour of simulated annealing, *Adv. Appl. Probability* Vol. 18, p. 747-771

[Paaß 89] Paaß, G. (1989): Stochastic generation of Synthetic Samples from Marginal Information. Proc. Fifth Annual Research Conference, Bureau of the Census, Washington D.C. p.431-445.

[Paaß 90] Paaß, G. (1990): Default Reasoning, Uncertain Reasoning, and Simulated Annealing. Tasso-Report No. 8, Aug. 1990, GMD, D-5205 St.Augustin, Germany.

[Paaß 91] Paaß, G. (1991): Probabilistic Default Reasoning. in: B.Bouchon-Meunier, R. Yager, L.A. Zadeh (eds.): Uncertainty in Knowledge Bases. Springer Verlag, p. 76-85.

[Paaß 91a] Paaß, G. (1991): Second order Probabilities for Uncertain and Conflicting Evidence. in: P.P. Bonissone, M. Henrion, L.N. Kanal, J.F. Lemmer

(eds.) Uncertainty in Artificial Intelligence 6, Elsevier Science Publishers, Amsterdam, pp. 447-456.

[Pearl 88] Pearl, J. (1988): *Probabilistic Reasoning in Intelligent Systems*, Morgan Kaufmann, San Mateo, Cal.

[Shore & Johnson 80] Shore, J.E., Johnson, R.W. (1980), Axiomatic Derivation of the Principle of Maximum Entropy and the Principle of Minimum Cross- Entropy, IEEE-Transactions on Information Theory, IT-26, 26-37.

## 8 APPENDIX

**Lemma 2 (Properties of $H$ for continuous $C$)**
We assume that $C(p)$ is bounded from below. Then $C^* = \inf_{p \in \mathcal{P}} C(p)$ exists. For $\delta \geq 0$ we define $D_\delta := \{p \in \mathcal{P} \mid C(p) \leq C^* + \delta\}$. We assume for some $\delta_0 > 0$ the function $C(p)$ is continuous in $D_{\delta_0}$. Then

$$H(p) \text{ has a maximum value } H^* \text{ in } D_0. \quad (18)$$

Let $V_0 := \{p \in \mathcal{P} \mid H(p) = H^*\}$ and $V_\epsilon := \{p \in \mathcal{P} \mid \exists_{p^* \in V_0} \forall_i \mid p_i - p_i^* \mid \leq \epsilon\}$ for $\epsilon > 0$. Let $D_\delta^\epsilon$ be the closure of $D_\delta \cap (\mathcal{P} \setminus V_\epsilon)$. Then for every $\epsilon > 0$ the following statements hold: There exists a $\Delta_{H^*}^\epsilon > 0$ with

$$\Delta_{H^*}^\epsilon = \left(H(p^*) - \max_{p \in D_0^\epsilon} H(p)\right)/3 \quad (19)$$

There is a $\delta^\epsilon \in (\epsilon, 0)$ such that for all $0 \leq \delta \leq \delta^\epsilon$ we have

$$\max_{p \in D_\delta^\epsilon} H(p) \leq H^* - 2\Delta_{H^*}^\epsilon \quad (20)$$

$$\max_{p \in D_\delta^\epsilon} C(p) \leq C^* + \delta^\epsilon \quad (21)$$

There is a $\eta \in (\epsilon, 0)$ such that for all $p \in V_\eta$

$$H(p) \geq H^* - \Delta_{H^*}^\epsilon \qquad C(p) \leq C^* + \delta^\epsilon/2 \quad (22)$$

If $p \in \mathcal{P} \setminus (V_\epsilon \cup D_{\delta^\epsilon}^\epsilon)$, then

$$C(p) > C^* + \delta^\epsilon \qquad H(p) \leq \log(m) \quad (23)$$

**Proof:** For all $\delta \in [0, \delta_0]$ the sets $D_\delta$ are closed as by the continuity of $C$ the limit of arbitrary sequences $\lim_{i \to \infty} p(i)$ is contained in $D_\delta$ if all $p(i) \in D_\delta$. Because of $D_\delta \subset \mathcal{P}$ they are in addition compact. Hence the continuous function $H(p)$ has an unique maximum value $H^*$ in $D_0$ which yields (18).

The sets $D_\delta^\epsilon$ are closed by construction. Hence $H$ takes a unique maximum value $H_0^\epsilon := \max_{p \in D_0^\epsilon} H(p)$. For each $p \in D_0^\epsilon$ we have by definition $\forall_i \mid p_i - p_i^* \mid > \epsilon$ if $H(p^*) = H^*$. As $H$ is continuous the maximum in $D_0^\epsilon$ has to be smaller than $H^*$.

$$H_0^\epsilon < \max_{p \in D_0} H(p) = H^*$$

Defining $\Delta_{H^*}^\epsilon := (H^* - H_0^\epsilon)/3 > 0$ yields (19).

The correspondence $F_1 : \delta \to D_\delta^\epsilon$ is continuous and compact-valued. Because of the continuity of $H$ the correspondence $f_1 : \delta \to \max_{p \in D_\delta^\epsilon} H(p)$ is a continuous function according to the maximum theorem [Green & Heller 81, p.49]. Because of $f_1(0) = H_0^\epsilon$ there is a $\delta^\epsilon > 0$, such that for all $\delta \leq \delta^\epsilon$ relation (21) holds. The inequality (21) follows directly from the definition of $D_\delta$.

The correspondence $F_2 : \eta \to V_\eta$ is continuous and compact-valued. By the continuity of $H$ the maximum theorem again yields that $f_2 : \epsilon \to \max_{p \in V_\epsilon} (-H(p))$ is a continuous function. Because of $f_2(0) = H(p^*)$ there is an $\epsilon_2$, such that for all $\tilde{\epsilon} \leq \epsilon_2$

$$\min_{p \in V_\epsilon} H(p) \geq H(p^*) - \Delta_{H^*}^\epsilon. \quad (24)$$

Because of the continuity of $C(p)$ there is an $\epsilon_3$, such that $C(p) \leq C^* + \delta^\epsilon/2$ for all $p \in V_{\epsilon_3}$. Setting $\eta = \min(\epsilon_2, \epsilon_3)$ yields (22). The relations (23) follow from the definition of $D_\delta^\epsilon$ and because $\max_{p \in S} H(p) = \log(m)$. ∎

**Proof of Theorem 1**

**Proof:** For $q \in \mathcal{Q}_n$ let $N_n(q)$ denote the number of different samples $X$ with identical counts $nq$. Csizár and Körner [Csiszár & Körner 81, p.30] show that

$$N_n(q) = \delta(q, n) \exp(nH(q)) \quad (25)$$

with $1/(n+1)^m \leq \delta(q, n) \leq 1$. By (7) we get the following equilibrium distribution for control parameter $\beta$:

$$\Pr(q) = N_n(q) c_0(\beta) \exp(-C(q)\beta)$$

This leads to the bounds

$$\Pr(q) \geq \frac{1}{(n+1)^m} c_0(\beta) \exp(nH(q) - C(q)\beta)$$

$$\Pr(q) \leq c_0(\beta) \exp(nH(q) - C(q)\beta)$$

Let $M^-$ be some subset of $\mathcal{Q}_n$ an $q^+ \in \mathcal{Q}_n \setminus M^-$. Note that $\mathcal{Q}_n$ contains at most $(n+1)^m$ elements. For the stationary distribution we get

$$\frac{\Pr(M^-)}{\Pr(q^+)} = \frac{\sum_{q \in M^-} \Pr(q)}{\Pr(q^+)}$$

$$\leq (n+1)^{2m} \frac{\exp(nH^{max}(M^-) - C^{min}(M^-)\beta)}{\exp(nH(q^+) - C(q^+)\beta)}$$

$$\leq (n+1)^{2m} \exp([H^{max}(M^-) - H(q^+)]n - [C^{min}(M^-) - C(q^+)]\beta) \quad (26)$$

where $C^{min}(M^-) := \inf_{q \in M^-} C(q)$ and $H^{max}(M^-) := \sup_{q \in M^-} H(q)$. $\Pr(M^-)/\Pr(q^+)$ has the general form $n^r \exp(-n^b)$. It is wellknown that $\lim_{n \to \infty} n^r \exp(-n^b) = 0$ for $r > 0$, $b > 0$. Hence a sufficient condition that $\Pr(M^-)/\Pr(q^+)$ converges to zero with growing $n$ is the existence of some $a, b > 0$ such that the exponent is smaller than $-an^b$.

Assume a fixed $\epsilon > 0$ is selected. Let us first define $M^- := D_{\delta^\epsilon}^\epsilon$ and $q_n^+$ as an element of $V_\eta \cap \mathcal{Q}_n$ with minimum cost $C_n^+$. Then we may for every $p \in \mathcal{P}$ find a $q \in \mathcal{Q}_n$ such that for all $i$ we have $\mid p_i - q_i \mid \leq \frac{1}{n}$. By the continuity of $C$ we have $\lim_{n\to\infty} C_n^+ = C^*$. Then with (21), (21), and (22) we get the following exponent $E_1(n)$ of (26):

$$\begin{aligned} E_1(n) &\leq [(H^* - 2\Delta_{H^*}^\epsilon) - (H^* - \Delta_{H^*}^\epsilon)]n \\ &\quad -[C^* - C_n^+]\beta \\ &\leq -\Delta_{H^*}^\epsilon n + [C_n^+ - C^*]\beta \end{aligned} \quad (27)$$

Let us now define $M^- := \mathcal{P} \setminus (V_\epsilon \cup D_{\delta^\epsilon}^\epsilon)$ and $q_n^+$ as before. Then from (22) and (23) we get the following exponent $E_2(n)$ of (26):

$$\begin{aligned} E_2(n) &= [\log(m) - (H^* - \Delta_{H^*}^\epsilon)]n \\ &\quad -[(C^* + \delta^\epsilon) - (C^* + \delta^\epsilon/2)]\beta \\ &\leq \log(m)n - \beta\delta^\epsilon/2 \end{aligned} \quad (28)$$

The ratio $\Pr^\circ(M^-)/\Pr^\circ(p^+)$ converges to 0 for growing $n$ if there are constants $a_i, b_i > 0$ such that $E_i(n) \leq a_i n^{b_i}$. Let us define $\beta_n := a_\beta n^{b_\beta}$. Assume in addition that there are $a_C, b_C > 0$ such that $C_n^+ - C^* \leq a_C n^{-b_C}$. Then we get from (27) the requirement $b_\beta < 1 + b_C$ and from (28) we get $b_\beta > 1$ yielding

$$0 < 1 < b_\beta < 1 + b_C \quad (29)$$

as a sufficient condition that with growing $n$ and $\beta_n$ defined as above the ratios $\Pr(M^-)/\Pr(q_n^+)$ converge to 0.

To prove (15) consider $p^*, p \in \mathcal{P}$ with cost value $C(p^*) = C^*$. As $C$ is continuously differentiable in $D_\delta$ the partial derivatives are bounded because $D_\delta$ is compact. By Taylors theorem there is a constant $a_C > 0$ such that $|C(p^*) - C(p)| \leq a_C \max_i |p_i^* - p_i|$ in $D_\delta$. If $C$ is twice continuously differentiable in $D_\delta$ then Taylors theorem implies that there is a constant $a_C > 0$ such that $|C(p^*) - C(p)| \leq a_C (\max_i |p_i^* - p_i|)^2$ in $D_\delta$ because the first derivative has to be zero in the minimum. As there is always a $q \in \mathcal{Q}_n$ such that $\max_i |p_i^* - q_i| \leq \frac{1}{n}$ the difference is bounded from above in both cases by $a_C n^{-b_C}$ with $b_C > 0$. ∎